# A Data-Efficient Analytical Prior Machine Learning Framework for Sound Reduction Frequency Prediction in Helmholtz Resonators

Jiaming Li*

School of Informatics, Computing, and Cyber Systems, Northern Arizona University, Flagstaff, Arizona 86011, USA

**Abstract:** High-fidelity finite-element simulations provide accurate numerical predictions of sound reduction frequencies for side-branch Helmholtz resonators, but large simulation datasets are expensive to generate and purely data-driven surrogates can become unreliable when simulation data are scarce. This study develops an analytical-prior learning framework that uses a low-cost analytical model to improve the data-efficient use of scarce high-fidelity simulation data. Two complementary routes are considered. When the analytical model remains available during prediction, it is retained as an explicit baseline and the limited simulation data are used to learn only the remaining analytical-to-simulation discrepancy. When a self-contained learned predictor is required, the analytical mapping is first distilled from abundant low-cost analytical evaluations into a learned prior and then calibrated using the limited simulation data. The framework was evaluated using rectangular side-branch Helmholtz resonators with 86 simulation-labelled geometries containing paired analytical and COMSOL-simulated sound reduction frequencies and 8,998 non-overlapping analytical-only geometries. The analytical model achieved a mean absolute error (MAE) of 1.333 Hz. Direct support vector regression (SVR) produced an MAE of 3.375 Hz, whereas analytical residual correction reduced the error to 0.426 Hz. A direct multilayer perceptron (MLP) trained from scratch achieved an MAE of 1.109 Hz. Analytical-prior pretraining reduced the error to 0.556 Hz with frozen-prior residual adaptation and to 0.371 Hz with full-model fine-tuning. Matched analyses using 20 to 70 simulation-labelled training cases further showed that both analytical correction and analytical-prior pretraining consistently improved data efficiency relative to their direct-learning counterparts. These results demonstrate that analytical prior information can enable scarce and expensive simulation data to be used more efficiently for accurate prediction of the high-fidelity numerical response, with explicit correction and prior distillation providing complementary solutions for different deployment requirements.

## 1. Introduction

Side-branch and Helmholtz-type resonators are widely used for passive control of tonal and narrow-band duct noise. Their sound-reduction frequency depends on neck inertia, cavity compliance, main-duct coupling, and geometric configuration. Analytical models provide rapid prediction and clear physical interpretation. Finite-element simulations can capture geometric and wave effects with higher fidelity, but large parameter sweeps require substantially greater computational effort [1–11]. Recent mass-spring analogy models have extended analytical prediction to asymmetric cavities and multiple openings [12–14]. These models capture much of the dominant geometry-frequency relationship, although systematic discrepancies from finite-element predictions remain.

Machine learning has become increasingly useful for acoustic surrogate modelling, inverse design, optimization, and resonator design [15–20]. A trained surrogate can provide rapid predictions and support large-scale design exploration. Conventional data-driven models usually learn the complete mapping from geometry to the target response. Their accuracy therefore depends on the quantity and coverage of the training data. This dependence becomes restrictive when the labels are generated by expensive finite-element simulations. Under a limited simulation budget, the available data must capture both the dominant physical trend and the remaining high-fidelity corrections. Part of this learning task can be redundant when a validated analytical model already represents much of the dominant response.

*Correspondence to: Jiaming Li, School of Informatics, Computing, and Cyber Systems, Northern Arizona University, Flagstaff, Arizona 86011, USA. E-mail: Jiaming.li@nau.edu

Multi-fidelity learning provides a natural basis for reducing this redundancy. Existing studies have combined abundant low-cost information with smaller high-fidelity datasets through information fusion, transfer learning, and residual correction [21–25]. Related approaches have been applied in computational mechanics, structural dynamics, and acoustic-wave prediction. Physics-guided machine learning further shows that prior physical knowledge can be introduced through training data, model structure, or learning objectives [26,27]. These studies establish the value of lower-fidelity information. However, the efficient reuse of a validated analytical resonator model under a fixed and limited high-fidelity simulation budget remains insufficiently studied. In this setting, the key objective is to use expensive simulation data mainly to learn information that is missing from the analytical model. This can be achieved by retaining the analytical prediction and learning only its discrepancy from simulation, or by first distilling the analytical relationship into a learned prior and then calibrating it with limited high-fidelity simulation data.

The present study develops an analytical-prior learning framework for the data-efficient use of scarce high-fidelity simulation data. The framework considers two complementary routes according to whether the analytical model remains available during prediction. When it is retained in the prediction pipeline, its output serves as an explicit baseline and the limited simulation data are used to learn only the remaining analytical-to-simulation discrepancy. When the final predictor is required to operate independently of the analytical model, the analytical mapping is first distilled from abundant low-cost analytical evaluations into a learned prior. This prior is then calibrated with the limited simulation data through either full-model fine-tuning or frozen-prior residual adaptation. Direct high-fidelity learning is used as a reference to quantify the improvement in data efficiency provided by the analytical prior.

The framework is evaluated using rectangular side-branch Helmholtz resonators. The case study combines 8,998 analytical-only geometries with 86 simulation-labelled geometries containing paired analytical and COMSOL frequencies. Residual support vector regression (SVR) is used to evaluate the explicit-prior route, while compact multilayer perceptrons (MLPs) are used to evaluate analytical-prior distillation and subsequent high-fidelity calibration. Matched experiments with progressively smaller simulation-labelled training sets further assess the data efficiency of both routes. The resulting comparison evaluates whether analytical prior information enables scarce and expensive simulation data to be used more efficiently for accurate high-fidelity prediction.

## 2. Methodology

The proposed analytical-prior learning framework addresses regression problems in which high-fidelity simulation data are expensive and scarce while a lower-cost analytical model is available for the same design variables and target quantity. The objective is to use the limited simulation-labelled data efficiently by directing high-fidelity learning toward information that is not already represented by the analytical model. Two complementary routes are defined according to the intended prediction setting. The analytical model can remain explicitly in the prediction pipeline and provide the baseline for residual correction, or its input-to-response relationship can be distilled into a learned prior before high-fidelity calibration. Direct high-fidelity learning is retained as a reference for evaluating the benefit of analytical prior information.

Let $\mathbf{x} \in \mathbb{R}^d$ denote the design variables, $A(\mathbf{x})$ the analytical prediction, and $S(\mathbf{x})$ the corresponding high-fidelity simulation response. For compact notation, $A_i = A(\mathbf{x}_i)$ and $S_j = S(\mathbf{x}_j)$ denote responses evaluated at individual design points. The analytical response at each simulation-labelled design point is denoted analogously. Two datasets are considered. The analytical-only dataset is $\mathcal{D}_A = \{(\mathbf{x}_i, A_i)\}_{i=1}^{N_A}$, whereas the simulation-labelled dataset is paired and is written as $\mathcal{D}_S = \{(\mathbf{x}_j, A_j, S_j)\}_{j=1}^{N_S}$. Each simulation-labelled case therefore contains the design variables, the analytical prediction, and the high-fidelity simulation response at the same geometry. The intended regime is $N_A \gg N_S$, because analytical evaluations can be generated at substantially lower cost than high-fidelity simulations. The framework assumes that the two fidelity levels share the same input variables and target quantity and that $A(\mathbf{x})$ is informative but imperfect. A generic regression loss is denoted by $\mathcal{L}$; trainable model parameters are introduced below as needed.

A direct learner provides the high-fidelity-only reference. Let $F_\theta(\mathbf{x})$ denote a parametric predictor with trainable parameters $\theta$. It uses only the pairs $(\mathbf{x}_j, S_j)$ from $\mathcal{D}_S$ and deliberately ignores the available analytical prediction. The parameters and resulting direct prediction are defined by

$$\theta_{\mathrm{dir}} = \underset{\theta}{\operatorname{argmin}} \frac{1}{N_S} \sum_{j=1}^{N_S} \mathcal{L}\left(F_\theta(\mathbf{x}_j), S_j\right) \tag{1}$$

The resulting direct prediction is $\hat{S}_{\mathrm{dir}}(\mathbf{x}) = F_{\theta_{\mathrm{dir}}}(\mathbf{x})$. This reference places the complete geometry-to-high-fidelity mapping on the scarce simulation-labelled data and provides a matched basis for assessing whether the analytical prior improves data efficiency.

When the analytical model can remain available during prediction, the high-fidelity target can be reformulated as a correction to the analytical baseline. The analytical-to-simulation discrepancy for each paired case is

$$\Delta_{A,j} = S_j - A_j \tag{2}$$

Let $R_\phi(\mathbf{x}, A(\mathbf{x}))$ denote a residual model with trainable parameters $\phi$. The residual model uses both the design variables and the analytical prediction and is fitted to the discrepancy targets by

$$\phi_A = \underset{\phi}{\operatorname{argmin}} \frac{1}{N_S} \sum_{j=1}^{N_S} \mathcal{L}\left(R_\phi(\mathbf{x}_j, A_j), \Delta_{A,j}\right) \tag{3}$$

The explicit-prior prediction is then reconstructed as

$$\hat{S}_{\mathrm{exp}}(\mathbf{x}) = A(\mathbf{x}) + R_{\phi_A}(\mathbf{x}, A(\mathbf{x})) \tag{4}$$

This formulation leaves the analytical prediction unchanged as the baseline and assigns the simulation-labelled data only to the remaining analytical-to-simulation correction. It does not require a separate large analytical-only training set. The analytical model must, however, still be evaluated for every new input at inference.

When the final predictor is required to operate independently of the analytical calculation, the analytical relationship is first distilled into model parameters. Let $P_\theta(\mathbf{x})$ denote a parametric predictor used for analytical-prior learning. The analytical-only dataset is used to estimate parameters that reproduce the analytical mapping:

$$\theta_A = \underset{\theta}{\operatorname{argmin}} \frac{1}{N_A} \sum_{i=1}^{N_A} \mathcal{L}\left(P_\theta(\mathbf{x}_i), A_i\right) \tag{5}$$

After this optimization, $P(\mathbf{x}) \equiv P_{\theta_A}(\mathbf{x})$ denotes the learned analytical prior and is expected to approximate $A(\mathbf{x})$ over the analytical training domain. The notation $P(\mathbf{x})$ is used for the frozen pretrained state, while $P_\theta(\mathbf{x})$ denotes the same model when its parameters remain trainable during calibration.

High-fidelity calibration of the distilled prior can follow two forms. In full-model fine-tuning, all pretrained parameters remain trainable. Starting from the analytical-prior parameters $\theta_A$, the model is optimized directly against the high-fidelity simulation response:

$$\theta_S = \underset{\theta}{\operatorname{argmin}} \frac{1}{N_S} \sum_{j=1}^{N_S} \mathcal{L}\left(P_\theta(\mathbf{x}_j), S_j\right) \tag{6}$$

The optimization is initialized from $\theta_A$. The calibrated prediction is $\hat{S}_{\text{full}}(\mathbf{x}) = P_{\theta_S}(\mathbf{x})$. The analytical mapping therefore provides the initial learned representation, while the limited simulation-labelled data are allowed to adjust the complete parameter set toward the high-fidelity response. The calibrated model is self-contained and no analytical evaluation is required at inference.

A more constrained calibration keeps the learned analytical prior fixed and uses the simulation-labelled data only to estimate the discrepancy that remains after distillation. Define $P_j = P(\mathbf{x}_j)$. The residual target is

$$\Delta_{P,j} = S_j - P_j \tag{7}$$

Let $R_\psi(\mathbf{x}, P)$ denote the corresponding residual calibrator with trainable parameters $\psi$. With the learned prior frozen, the calibration parameters are obtained from

$$\psi_P = \underset{\psi}{\operatorname{argmin}} \ \frac{1}{N_S} \sum_{j=1}^{N_S} \mathcal{L}\left(R_\psi(\mathbf{x}_j, P_j), \Delta_{P,j}\right) \tag{8}$$

and the final frozen-prior prediction is

$$\hat{S}_{\text{fr}}(\mathbf{x}) = P(\mathbf{x}) + R_{\psi_P}(\mathbf{x}, P(\mathbf{x})) \tag{9}$$

This form preserves the distilled analytical mapping and confines high-fidelity learning to an additive correction. Both full-model fine-tuning and frozen-prior residual adaptation produce self-contained learned predictors because the original analytical calculation is not required after the prior has been distilled.

The framework therefore defines two complementary uses of the same analytical information. Explicit analytical correction applies when the analytical model can remain in the prediction pipeline, allowing scarce simulation-labelled data to focus on its discrepancy. Analytical-prior distillation applies when a self-contained learned predictor is required. Within the distilled-prior route, full-model fine-tuning permits global adaptation of the learned prior, whereas frozen-prior residual adaptation preserves the prior and limits high-fidelity learning to the remaining correction. Direct high-fidelity learning serves as the common reference. Section 3 applies these formulations to rectangular side-branch Helmholtz resonators.

# 3. Case-study validation

## 3.1 Study system and problem formulation

The analytical-prior learning framework is evaluated using a family of rectangular side-branch Helmholtz resonators for which both analytical predictions and high-fidelity finite-element results are available. The resonator family includes variations in cavity dimensions, neck geometry, and cavity asymmetry, providing a controlled setting for evaluating the data efficiency of the two analytical-prior routes. The prediction target is the sound-reduction frequency of each resonator geometry. Each resonator is represented by six continuous geometric variables. The geometry vector is defined as

$$\mathbf{x} = [r,\ W,\ H,\ H_m,\ H_n,\ W_1]^{\mathrm{T}} \in \mathbb{R}^6 \tag{10}$$

where $r$ is neck radius; $W$ and $H$ are resonator width and height; $H_m$ is main-duct height; $H_n$ is neck height; and $W_1$ is left-cavity width measured relative to the neck centre. The final variable provides a continuous descriptor of cavity asymmetry.

For each geometry $\mathbf{x}$, the analytical model provides the low-cost prediction $A(\mathbf{x})$, while the finite-element simulation provides the high-fidelity response $S(\mathbf{x})$. The analytical predictions are obtained from equivalent mass-spring formulations covering symmetric, asymmetric, and multiple-opening resonators [12–14]. For the multi-opening cases, a cluster of closely spaced necks is represented by an equivalent single opening spanning the outer edges of the leftmost and rightmost necks, following the enhanced analytical formulation [14]. These formulations provide physically interpretable geometry-dependent predictions while remaining substantially less expensive to evaluate than finite-element simulations. Following the general framework in Section 2, the paired analytical-to-simulation discrepancy is $\Delta_A(\mathbf{x}) = S(\mathbf{x}) - A(\mathbf{x})$. The COMSOL response $S(\mathbf{x})$ serves as the high-fidelity numerical reference throughout the case study. Direct learning predicts $S(\mathbf{x})$ from the geometric variables alone, whereas the explicit-prior route uses the analytical prediction to learn the remaining discrepancy. The distilled-prior route uses additional analytical-only cases to learn the analytical mapping before calibration with the same simulation-labelled data.

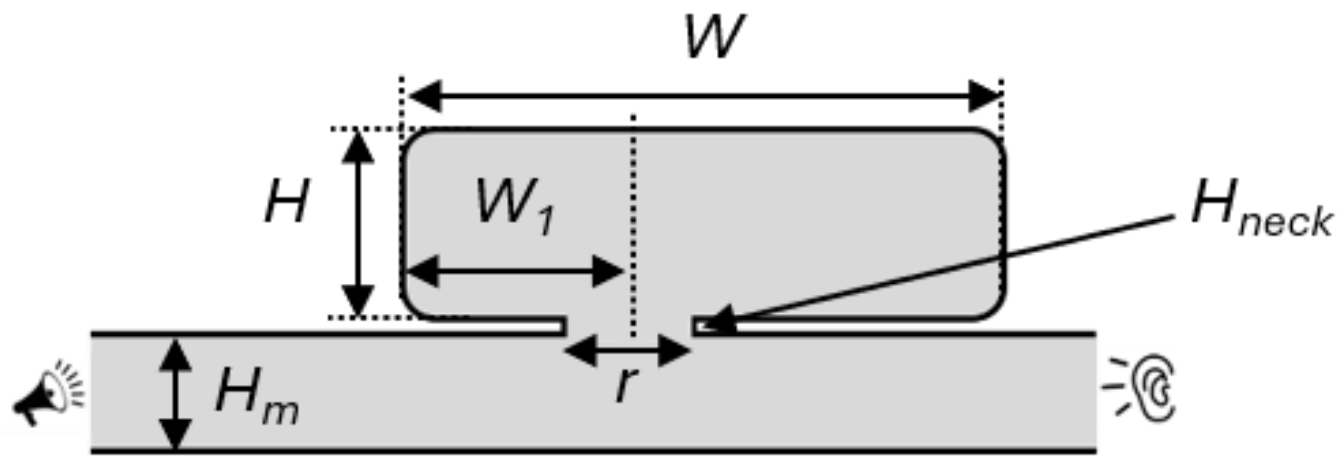

*Figure 1. Rectangular side-branch resonator and geometric variables used in the study. $H_m$ denotes main-duct height; $W$ and $H$ denote resonator width and height; $r$ is neck radius; $H_n$ is neck height; and $W_1$ is left-cavity width. All dimensions are measured in metres.*

### 3.2 Analytical and simulation-labelled datasets

The case study combines a large analytical pool with a smaller simulation-labelled dataset. The original analytical pool contains 9,000 model-generated cases, comprising 3,000 symmetric single-neck, 3,000 asymmetric single-neck, and 3,000 multiple-opening geometries. The simulation-labelled dataset contains 86 geometries with paired analytical and COMSOL frequencies, comprising 21 symmetric single-neck, 24 asymmetric single-neck, and 41 multiple-opening cases.

Two analytical cases that exactly matched simulation-labelled geometries across all six geometric variables were removed before analytical-prior distillation. This left 8,998 non-overlapping analytical-only cases. Structural groups were used only for stratification and were not included as predictive inputs.

The analytical pool spans 0.010–0.650 m in neck radius, 1.050–2.450 m in resonator width, 0.220–0.680 m in resonator height, 0.120–0.280 m in main-duct height, 0.002–0.028 m in neck height, and 0.100–2.300 m in left-cavity width. Within the 86 simulation-labelled cases, the corresponding ranges are 0.030–0.120, 1.100–1.900, 0.280–0.440, 0.110–0.190, 0.001–0.003, and 0.150–1.350 m, respectively.

The high-fidelity frequencies were inherited from the COMSOL validation datasets used in the analytical-model studies [12–14]. They were generated in COMSOL Multiphysics 5.4 using the frequency-domain Pressure Acoustics interface under linear, no-mean-flow conditions, with a harmonic plane-wave inlet, a plane-wave radiation outlet, and sound-hard rigid walls; thermoviscous losses were neglected to remain consistent with the analytical assumptions. The sound-reduction frequency was identified as the frequency corresponding to maximum transmission loss.

Analytical frequencies span 21.50–268.86 Hz in the original 9,000-case pool. Within the 86 simulation-labelled cases, analytical predictions span 50.15–107.98 Hz and COMSOL frequencies span 51.10–114.20 Hz.

### 3.3 Model implementation and training

The case study evaluates three analytical-prior implementations under matched high-fidelity data budgets. The explicit-prior route retains the analytical prediction $A(\mathbf{x})$ during inference and is implemented using residual support vector regression (SVR). The distilled-prior route first learns a neural approximation $P(\mathbf{x})$ of the analytical mapping from the 8,998 analytical-only cases and then calibrates the learned prior using the 86 simulation-labelled cases through either full-model fine-tuning or frozen-prior residual adaptation. Their performance is assessed against three reference formulations: the original analytical model as a physics-only reference, direct SVR as the matched high-fidelity-only reference for residual SVR, and direct multilayer perceptron (MLP) as the matched high-fidelity-only neural reference for the two distilled-prior calibrations.

All simulation-based models use the same repeated stratified fivefold outer partitions with five repeats, ensuring matched training and test data across comparisons. The overall case-study implementation is summarized in Figure 2.

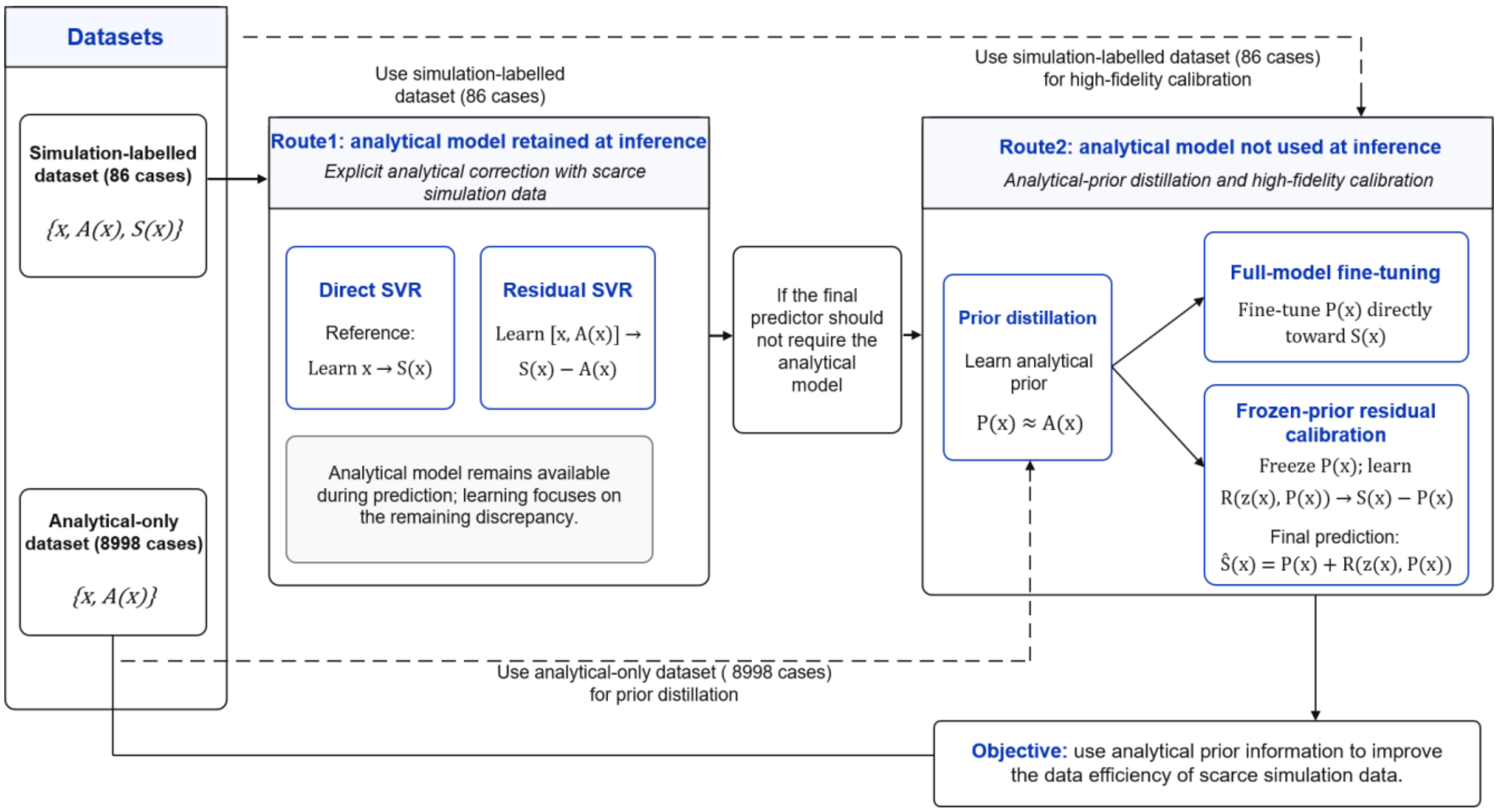

*Figure 2. Case-study validation framework for analytical-prior learning. The 86 simulation-labelled cases are used for direct learning, explicit analytical correction, and high-fidelity calibration, whereas the 8,998 analytical-only cases are used only for analytical-prior distillation. Route 1 retains the analytical model during inference, and Route 2 uses a distilled analytical prior followed by high-fidelity calibration.*

For the explicit-prior route, direct and residual SVR are trained on identical outer folds. Direct SVR uses the six geometric variables $\mathbf{x}$ to predict the high-fidelity response $S(x)$. Residual SVR instead uses the augmented input $[x, A(x)]$ and predicts the analytical-to-simulation discrepancy $S(x) - A(x)$; its final frequency estimate is obtained by adding the predicted correction to $A(x)$. Both formulations use a radial-basis-function kernel and standardized input features. Within each outer-training fold, fourfold stratified inner cross-validation selects $C \in \{1, 10, 100\}$ and $\varepsilon \in \{0.05, 0.2, 0.5\}$ by minimum validation mean absolute error (MAE), while $\gamma$ uses the scikit-learn scale rule. Feature standardization, hyperparameter selection, and fitting are confined to the corresponding training data.

For the distilled-prior route, a compact 6–64–32–1 MLP with rectified linear unit activations is first trained to reproduce the analytical response, such that $P(x) \approx A(x)$. A stratified 90/10 split of the 8,998 analytical-only cases is used only to select the pretraining duration. The split contains 8,098 training cases and 900 validation cases and selected 292 epochs in the reported run. A final analytical prior is then trained from scratch on all 8,998 analytical-only cases for the selected 292 epochs. Pretraining uses mean-squared error with AdamW, a learning rate of $10^{-3}$, weight decay of $10^{-4}$, and batch size 256. Input and analytical-target standardization are fitted on the analytical-only data used for the final prior.

A direct MLP with the same 6–64–32–1 architecture is trained from scratch on simulation-labelled data as the neural high-fidelity-only reference. Within each outer-training fold, a stratified 20% subset is used to select the training duration; a fresh model is then refitted on the complete outer-training fold for the selected number of epochs. Direct MLP training uses mean-squared error with AdamW, a learning rate of $10^{-3}$, weight decay of $10^{-4}$, batch size 16, a maximum epoch-selection period of 350 epochs, and patience 35. Input and simulation-target scalers are fitted using the corresponding outer-training data only. The same within-fold epoch-selection and full-fold refitting procedure is applied to both high-fidelity calibration strategies described below.

High-fidelity calibration of the distilled prior is evaluated in two forms. In full-model fine-tuning, the complete 6–64–32–1 network is initialized from the analytical-prior state and all 2,561 model parameters remain trainable while the model is optimized directly toward $S(x)$. The analytical-pretraining input and target scalers are retained so that calibration begins from the distilled prior without changing its initial mapping. Fine-tuning uses AdamW with a learning rate of $3 \times 10^{-4}$, weight decay of $10^{-4}$, batch size 16, a maximum epoch-selection period of 220 epochs, and patience 30.

In frozen-prior residual adaptation, the complete pretrained MLP, including its analytical output head, remains fixed. For each geometry, the frozen network provides the 32-dimensional latent representation $z(x)$ from its final hidden layer together with the scalar prior prediction $P(x)$. A 33–16–1 residual adapter receives $[z(x), P(x)]$ and is trained to predict $S(x) - P(x)$. The adapter output is added to the frozen prior to obtain the final high-fidelity prediction:

$$\hat{S}_{res}^{MLP}(x) = P(x) + R_{\varphi}([z(x), P(x)]) \tag{11}$$

The adapter final layer is initialized to zero, so the initial calibrated prediction is exactly $P(x)$. The 2,561 parameters of the pretrained MLP remain frozen, and only the 561 parameters of the residual adapter are trainable during high-fidelity calibration. Adapter features are standardized using only the corresponding high-fidelity training data. Residual targets are scaled by their training-set standard deviation without mean subtraction, preserving the interpretation that zero adapter output corresponds to zero correction. Residual calibration uses AdamW with a learning rate of $10^{-3}$, weight decay of $10^{-4}$, batch size 16, a maximum epoch-selection period of 350 epochs, and patience 40.

This implementation keeps the experimental comparison aligned with the two deployment settings of the analytical-prior framework. Residual SVR evaluates explicit analytical correction when $A(x)$ remains available during prediction. The two pretrained MLP strategies evaluate self-contained predictors in which analytical information has first been distilled into $P(x)$. Within each model family, the direct learner uses the same simulation-labelled partitions and serves as the matched reference for quantifying the benefit of analytical prior information.

### 3.4 Evaluation protocol and small-data robustness analysis

Because only 86 simulation-labelled cases are available, predictive performance is evaluated using five repeats of stratified fivefold cross-validation. The three structural groups are used only for stratification, and all simulation-based methods share the same 25 outer train/test splits. Each geometry receives one out-of-fold (OOF) prediction in each repeat. The five OOF predictions for the same geometry are averaged to form one case-level repeated-OOF estimate, and headline metrics are calculated across the resulting 86 predictions.

The primary performance metric is mean absolute error (MAE) in hertz, with the coefficient of determination ($R^2$) reported as a complementary goodness-of-fit measure:

$$MAE = \frac{1}{N}\sum_{i=1}^{N}\left|\hat{S}_i - S_i\right| \tag{12}$$

$$R^2 = 1 - \frac{\sum_{i=1}^{N}\left(\hat{S}_i - S_i\right)^2}{\sum_{i=1}^{N}\left(S_i - \overline{S}\right)^2} \tag{13}$$

Root mean squared error (RMSE) and mean absolute percentage error (MAPE) are retained as secondary diagnostics. MAE is emphasized because it directly represents the average frequency-prediction error in hertz.

The repeated-OOF predictions are used only for comparative evaluation and do not represent a single final deployed model.

Uncertainty in the principal pairwise comparisons is assessed by paired case-level bootstrap resampling. Ten thousand bootstrap samples are drawn from the 86 cases using seed 2026. The MAE gain is defined as reference-method MAE minus candidate-method MAE, so a positive value favours the candidate. Bootstrap intervals and resampling frequencies are treated as descriptive measures of comparison stability rather than formal P values.

Small-data robustness is evaluated at simulation-training budgets from 20 to 70 cases in increments of 10, using 30 paired stratified train/test splits at each budget. Direct and residual SVR share the same split, as do the direct MLP and full-model calibrated MLP. SVR hyperparameters are fixed to the modal settings from the primary analysis, while MLP training durations are fixed to the corresponding median selected values of 197 and 158 epochs. Because the held-out test size changes with the training budget, comparisons are interpreted within each budget rather than as a formal learning curve across budgets.

# 4. Results

## 4.1 Controlled performance comparisons of the two analytical-prior routes

The explicit-prior route was evaluated by comparing residual SVR with the analytical model as a physics-only reference and direct SVR as the matched high-fidelity-only reference (Figure 3a). Residual SVR produced substantially lower error than both references, reducing MAE by 68.0% relative to the analytical model and by 87.4% relative to direct SVR.

The distilled-prior route was evaluated against direct MLP as the matched high-fidelity-only neural reference (Figure 3b). Both analytical-prior strategies improved on direct MLP. The frozen-prior residual MLP reduced MAE by 49.8%, while the full-model calibrated MLP reduced it by 66.6% and gave the lowest numerical MAE, 0.371 Hz, with $R^2$ = 0.998.

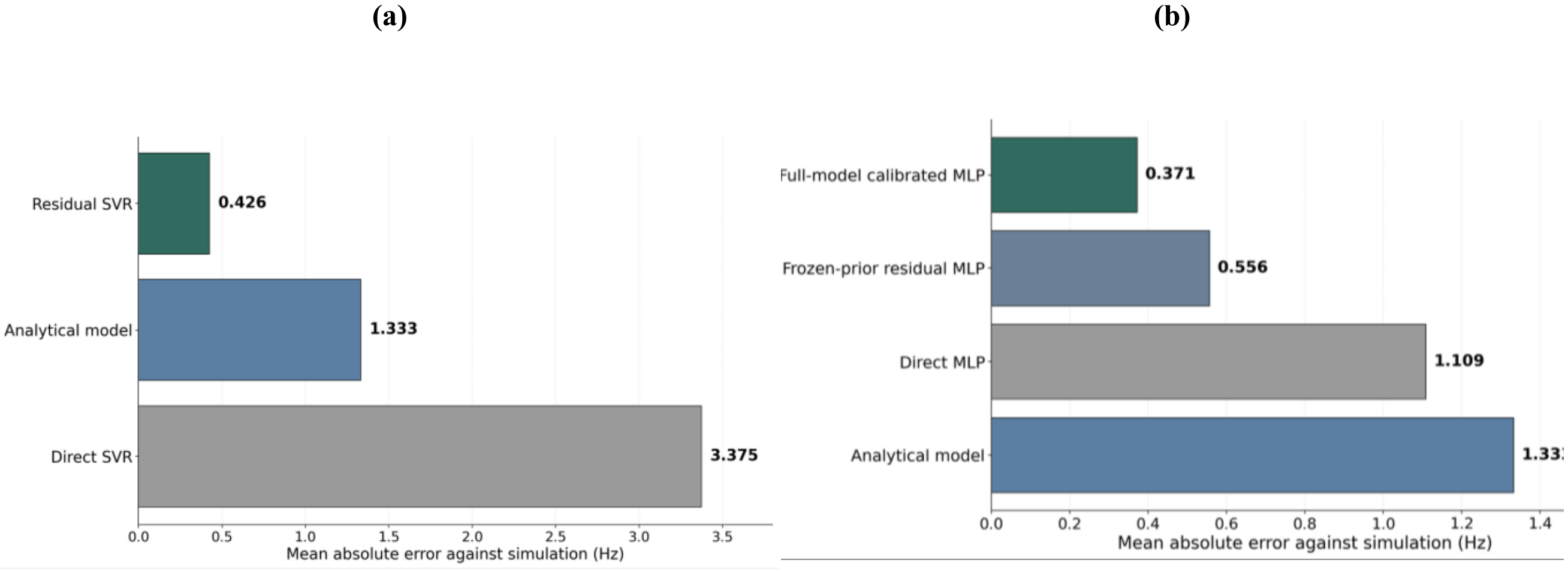

*Figure 3. Controlled performance comparisons for the two analytical-prior routes. (a) Explicit-prior route: residual SVR versus the analytical and direct-SVR references. (b) Distilled-prior route: frozen-prior residual and full-model calibrated MLPs versus the direct-MLP reference; the analytical model is also shown. Learned-model values use case-level repeated-OOF predictions, whereas the analytical reference is evaluated directly on the same 86 cases.*

**Table 1. Paired bootstrap comparisons of key prediction routes.**

| Candidate method | Reference method | MAE gain (Hz) | 95% bootstrap interval |
|---|---|---|---|
| Residual SVR | Analytical model | 0.907 | 0.737–1.078 |
| Residual SVR | Direct SVR | 2.950 | 2.034–4.011 |
| Full-model calibrated MLP | Direct MLP | 0.738 | 0.487–1.012 |
| Full-model calibrated MLP | Frozen-prior residual MLP | 0.185 | 0.079–0.293 |
| Full-model calibrated MLP | Residual SVR | 0.055 | −0.039–0.150 |

*Note: $MAE\ gain = MAE_reference - MAE_candidate$; positive values favour the candidate method. The interval for full-model calibrated MLP versus residual SVR crosses zero.*

Paired bootstrap comparisons confirmed the improvements over the matched direct-learning references (Table 1). Full-model fine-tuning also outperformed frozen-prior residual adaptation. In the descriptive cross-model comparison, the 0.055-Hz numerical difference between full-model calibrated MLP and residual SVR was small, and its 95% bootstrap interval crossed zero. The full-model calibrated MLP had lower MAE in 87.5% of bootstrap resamples, but the comparison was not decisive.

### 4.2 Fidelity of analytical-prior distillation and high-fidelity calibration

Before high-fidelity calibration, the distilled neural prior $P(x)$ was evaluated on the 86 simulation geometries that had been excluded from analytical pretraining. It reproduced the analytical prediction $A(x)$ with MAE = 0.499 Hz and $R^2$ = 0.997. Against the COMSOL target before calibration, the same prior had MAE = 1.342 Hz, close to the 1.333 Hz error of the exact analytical model. The pretrained network therefore recovered the analytical mapping closely while retaining a small distillation error.

After full-model high-fidelity calibration, the learned predictor closely tracked the COMSOL response $S(x)$ across most of the sampled frequency range, with MAE = 0.371 Hz and $R^2$ = 0.998. The largest visible deviation occurred at the upper end of the sampled frequency range.

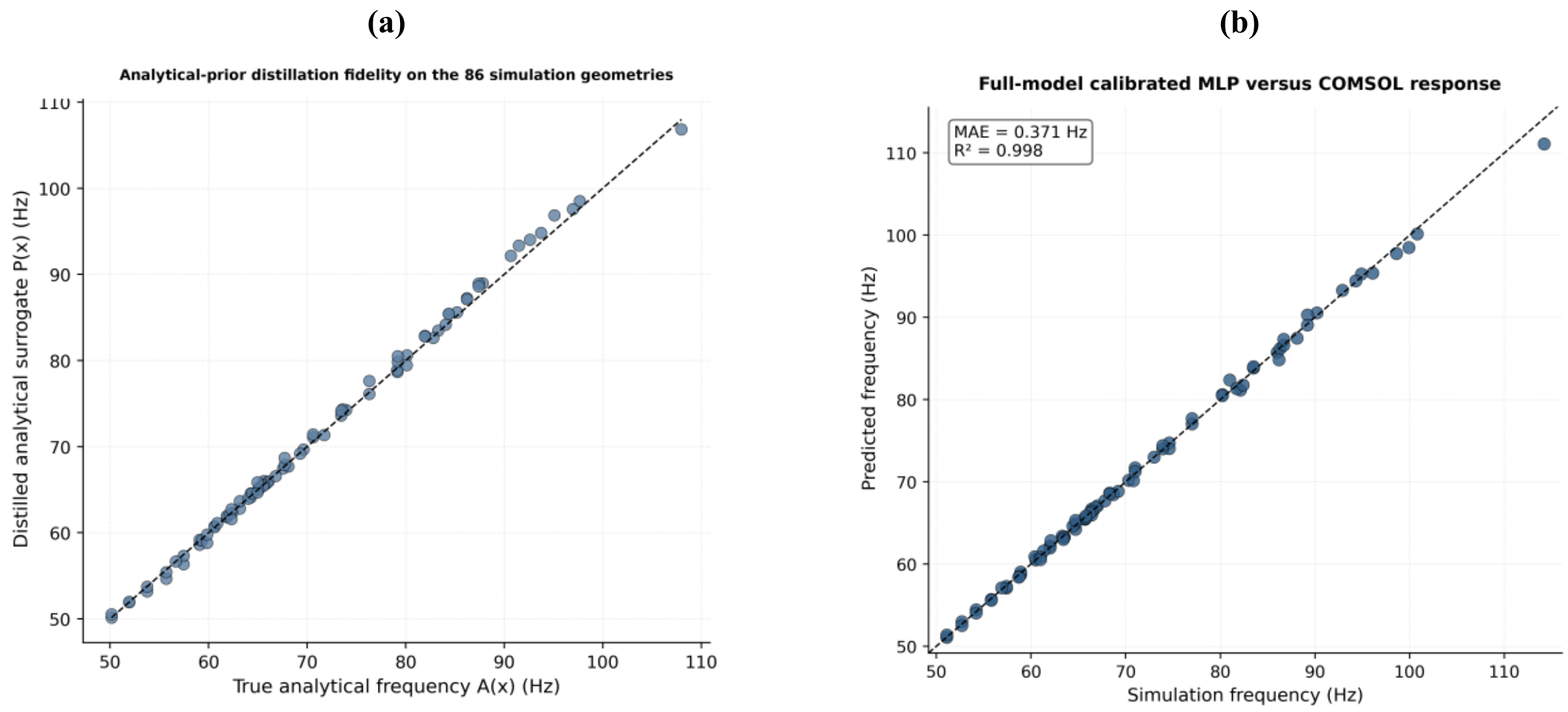

*Figure 4. Analytical-prior distillation and calibrated prediction. (a) Distilled neural prior $P(x)$ versus the analytical prediction $A(x)$ on the simulation geometries excluded from pretraining. (b) Case-level repeated out-of-fold predictions from the full-model calibrated MLP versus the COMSOL response $S(x)$. Dashed lines denote identity.*

### 4.3 Small-data robustness under reduced simulation budgets

The advantage of analytical prior information persisted as the simulation-labelled training budget was reduced from 70 to 20 cases (Figure 5). Residual SVR had lower mean test MAE than direct SVR at every tested budget and showed smaller between-split variability. With 20 training cases, mean MAE was 0.786 Hz for residual SVR and 4.448 Hz for direct SVR; with 70 cases, the corresponding values were 0.445 and 3.427 Hz.

The distilled-prior comparison showed the same pattern. With 20 simulation-labelled training cases, the full-model calibrated MLP achieved mean MAE = 0.846 Hz compared with 3.938 Hz for direct MLP. At 70 cases, the corresponding values were 0.339 and 1.105 Hz. Because the held-out test size changes with the training budget, these results are interpreted as matched within-budget comparisons rather than as a formal independent learning curve.

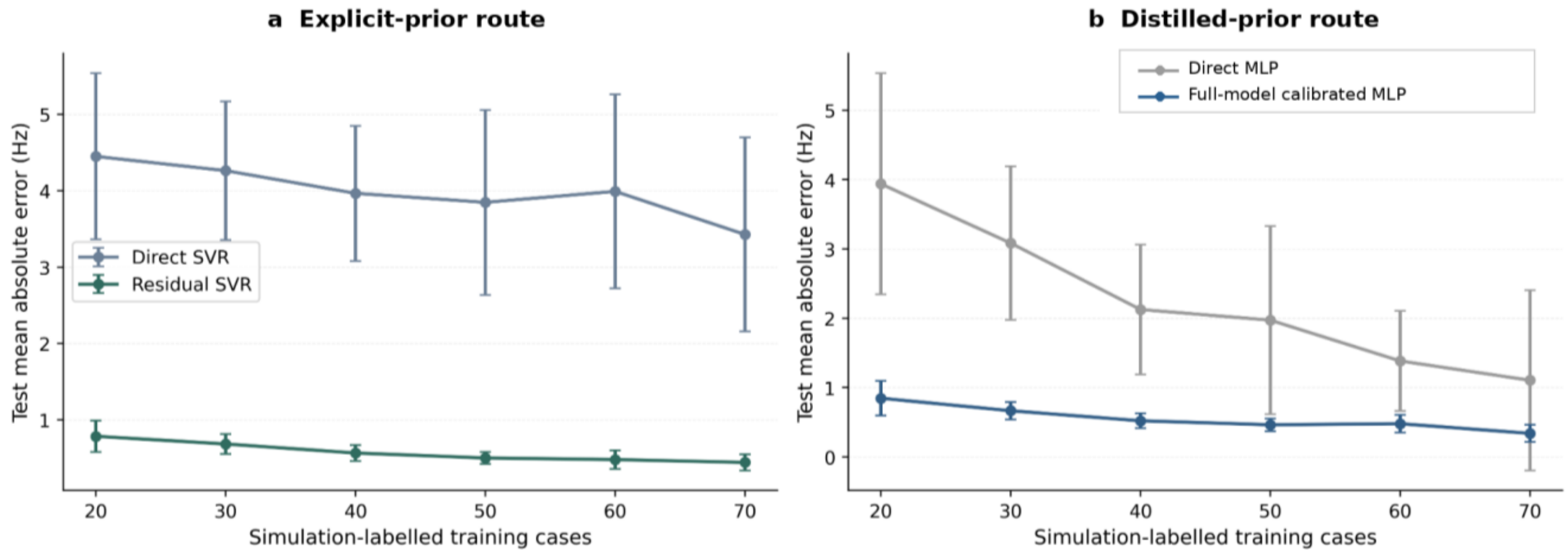

*Figure 5. Small-data robustness of the two analytical-prior routes. (a) Direct versus residual SVR. (b) Direct MLP versus full-model calibrated MLP. Points show mean test MAE across 30 paired stratified splits at each training budget; error bars show ±1 s.d. across splits.*

# 5. Discussion

## 5.1 Analytical priors improve the use of scarce high-fidelity data

The central finding is that scarce high-fidelity simulation data are used more effectively when information already available from a low-cost analytical model is incorporated into the learning process. The analytical model captures much of the dominant geometry-frequency relationship, whereas direct learners must reconstruct both this broad trend and the remaining analytical-to-simulation discrepancy from the same limited simulation-labelled dataset. Residual SVR reduces this burden by retaining the analytical prediction explicitly, while analytical-prior distillation shifts much of the initial learning to the larger analytical-only dataset.

The controlled comparisons support this interpretation across both model families. Residual SVR substantially outperformed direct SVR under matched high-fidelity folds, and both distilled-prior MLP calibrations improved on the direct MLP trained from scratch. The same pattern persisted as the simulation-training budget decreased from 70 to 20 cases. These results indicate that the benefit arises from how prior analytical information is used, rather than from access to additional high-fidelity labels.

## 5.2 Explicit correction and prior distillation serve different deployment needs

The two analytical-prior routes address different deployment settings. Explicit analytical correction is attractive when the analytical model can remain available at inference. In that setting, residual SVR retains the exact analytical prediction as the baseline and uses scarce simulation data only to learn the remaining correction. Analytical-prior distillation is more suitable when a self-contained predictor is preferred. The analytical mapping is first learned from inexpensive analytical evaluations and is then calibrated toward the high-fidelity response.

Within the distilled-prior route, full-model fine-tuning achieved lower error than frozen-prior residual adaptation. This difference is consistent with their adaptation constraints: full-model fine-tuning allows the complete pretrained representation to adjust toward the simulation response, whereas the frozen-prior route preserves the distilled prior and limits high-fidelity learning to an additive correction. The frozen route also operates around $P(x)$, an approximation of $A(x)$, so the correction must absorb both the analytical-to-simulation discrepancy and any remaining distillation error. Nevertheless, both calibrated MLPs clearly outperformed the direct MLP, showing that the distilled analytical prior remains useful under either adaptation strategy.

The two lowest-error approaches, residual SVR and full-model calibrated MLP, were numerically close and their paired bootstrap interval included zero. Their practical distinction is therefore more important than their ranking. Residual SVR preserves an explicit analytical component, whereas the calibrated MLP produces a single geometry-to-high-fidelity predictor after training. Route selection should consequently depend on the intended inference workflow as well as predictive accuracy.

### 5.3 Broader methodological significance

The framework is consistent with the broader objective of multi-fidelity learning: use abundant low-cost information to reduce dependence on expensive high-fidelity data [21–25]. The present setting is distinct in that the low-fidelity source is an established analytical acoustics model rather than another numerical solver. Analytical information enters either as an explicit prediction baseline or as supervised low-cost data used to form a learned prior, rather than as a governing-equation penalty, which distinguishes the present approach from conventional physics-informed neural networks [26,27].

The broader implication is that analytical, reduced-order, or empirical engineering models need not serve only as benchmarks for machine-learning surrogates. When such models already capture a meaningful part of the response, they can be used to reduce the amount of information that scarce simulation or experimental data must supply. The resonator case study provides a controlled demonstration of this principle through two complementary deployment routes and shows that their advantages remain evident under increasingly limited high-fidelity data budgets.

### 5.4 Limitations and future work

Several limitations define the scope of the present findings. First, COMSOL serves as the high-fidelity reference, so the reported accuracy reflects agreement with numerical simulation rather than laboratory measurements. Experimental validation is therefore needed, particularly under more realistic conditions such as thermoviscous losses, manufacturing tolerances, and mean flow.

Second, the simulation-labelled dataset contains only 86 geometries from the same rectangular resonator family. Repeated cross-validation improves the reliability of the evaluation, but it does not replace independent prospective testing or establish performance outside the sampled design space. Future work should therefore include new geometries and explicit applicability-domain assessment.

Third, the present study predicts a single sound-reduction frequency and evaluates a limited set of learning architectures. Extending the framework to full transmission-loss spectra, bandwidth, inverse design, target-frequency design [28], alternative calibration strategies, and uncertainty-aware prediction would help determine its broader applicability. Deployment would also require a fixed final-training procedure and external validation on previously unseen designs.

## 6. Conclusion

This study develops an analytical-prior learning framework for engineering prediction problems in which high-fidelity data are scarce but a low-cost analytical model is available. The framework provides two complementary routes for reusing analytical information. When the analytical model remains available during inference, it is retained as an explicit baseline and scarce high-fidelity data are used to learn only the remaining discrepancy. When a self-contained predictor is required, the analytical mapping is first distilled into a learned prior and then calibrated using the limited high-fidelity data.

The framework was validated using rectangular side-branch Helmholtz resonators with 8,998 analytical-only geometries and 86 simulation-labelled geometries. Residual SVR reduced MAE from 1.333 Hz for the analytical model and 3.375 Hz for direct SVR to 0.426 Hz. In the distilled-prior route, direct MLP achieved 1.109 Hz MAE, whereas frozen-prior residual adaptation and full-model fine-tuning reduced the error to 0.556 and 0.371 Hz, respectively. Both analytical-prior routes also retained clear advantages as the simulation-training budget decreased from 70 to 20 cases, demonstrating improved data efficiency in the low-data regime.

These results show that, when an analytical model already captures a meaningful part of the system response, scarce high-fidelity data can be used more efficiently for refinement and calibration than for reconstructing the complete mapping from scratch. The contribution therefore extends beyond the present resonator problem. Analytical, reduced-order, or empirical engineering models can serve as reusable prior information for data-efficient high-fidelity prediction, with explicit correction and prior distillation providing alternative implementations for different deployment requirements. The framework thus offers a general strategy for combining inexpensive existing models with limited and costly high-fidelity data in engineering surrogate modelling.

## Data and code availability

The data and code used in this study are available from the corresponding author upon reasonable request.

## Conflict of interests

The author declares no conflict of interests.